\documentclass[cameraready]{Interspeech}

\title{Contextual Earnings-22: A Speech Recognition Benchmark with Custom Vocabulary in the Wild}

\author[affiliation={1,2}, correspondingauthor]{Berkin}{Durmus}
\author[affiliation={1}]{Chen}{Cen}
\author[affiliation={1}]{Eduardo}{Pacheco}
\author[affiliation={1}]{Arda}{Okan}
\author[affiliation={1}]{Atila}{Orhon}

\address{
    $^1$ Argmax, Inc., USA \\
    $^2$ University of California, Los Angeles, USA
}

\email{\{berkin,chen,eduardo,arda,a\}@argmaxinc.com}


\renewcommand{\paragraph}[1]{\par\vspace{0.1\baselineskip}\noindent\textbf{#1}\ }

\usepackage{comment}
\usepackage{bookmark}
\usepackage{caption}
\usepackage{float}
\usepackage{tikz}
\usetikzlibrary{positioning,arrows.meta,shapes.geometric,calc}

\begin{document}

\maketitle

\begin{abstract}
  The accuracy frontier of speech-to-text systems has plateaued on academic benchmarks.\footnotemark[1] In contrast, industrial benchmarks and adoption in high-stakes domains suggest otherwise. We hypothesize that the primary difference between the two is contextual conditioning: Academic benchmarks are dominated by frequently encountered general vocabulary that is relatively easy to recognize compared with rare and context-defined custom vocabulary that has disproportionate impact on the usability of speech transcripts. Despite progress on contextual speech-to-text, there is no standardized benchmark. We introduce Contextual Earnings-22, an open dataset built upon Earnings-22, with realistic custom vocabulary contexts to foster research and reveal latent progress. We set six strong baselines for two dominant approaches: keyword prompting and keyword boosting. Experiments show both reach comparable and significantly improved accuracy when scaled from proof-of-concept to large-scale systems.\footnotemark[1]


\end{abstract}
\footnotetext[1]{The top-11 models on Hugging Face OpenASR Leaderboard \cite{hf_open_asr_leaderboard} as of December 2025 achieve very close average WER across 8 commonly used benchmark datasets. Part of this narrow gap is also attributed to verbatim vs non-verbatim behavior variance across models and may point to a near-saturation of WER improvements.}

\section{Introduction and Related Work}
Speech-to-text (STT) has reached high levels of accuracy on widely used academic benchmarks, to the point that reported word error rate (WER) improvements are often marginal across top-performing systems.\footnotemark[1] This apparent mismatch suggests that commonly reported benchmark WER may no longer be a sufficient proxy for real-world transcript utility.
A key driver of this mismatch is \emph{contextual conditioning}: in many applications, a small set of \emph{context-defined} (custom) terms disproportionately determines whether a transcript is usable.
In earnings calls, for example, a transcript can be otherwise fluent yet fail in practice if it repeatedly misrecognizes company, product, or person names.
This creates a regime where overall WER can be near-saturated while \emph{custom term accuracy} remains far from solved.
This paper studies the general problem of speech recognition with custom vocabulary (also called \emph{context biasing} in prior work \cite{ctcws,turbobias,flexctc,xu23d,le21,fox22,hou25}).
We use \emph{custom vocabulary} to emphasize the problem setting rather than a specific method.

\begin{figure}[!t]
  \centering
  \includegraphics[width=\linewidth]{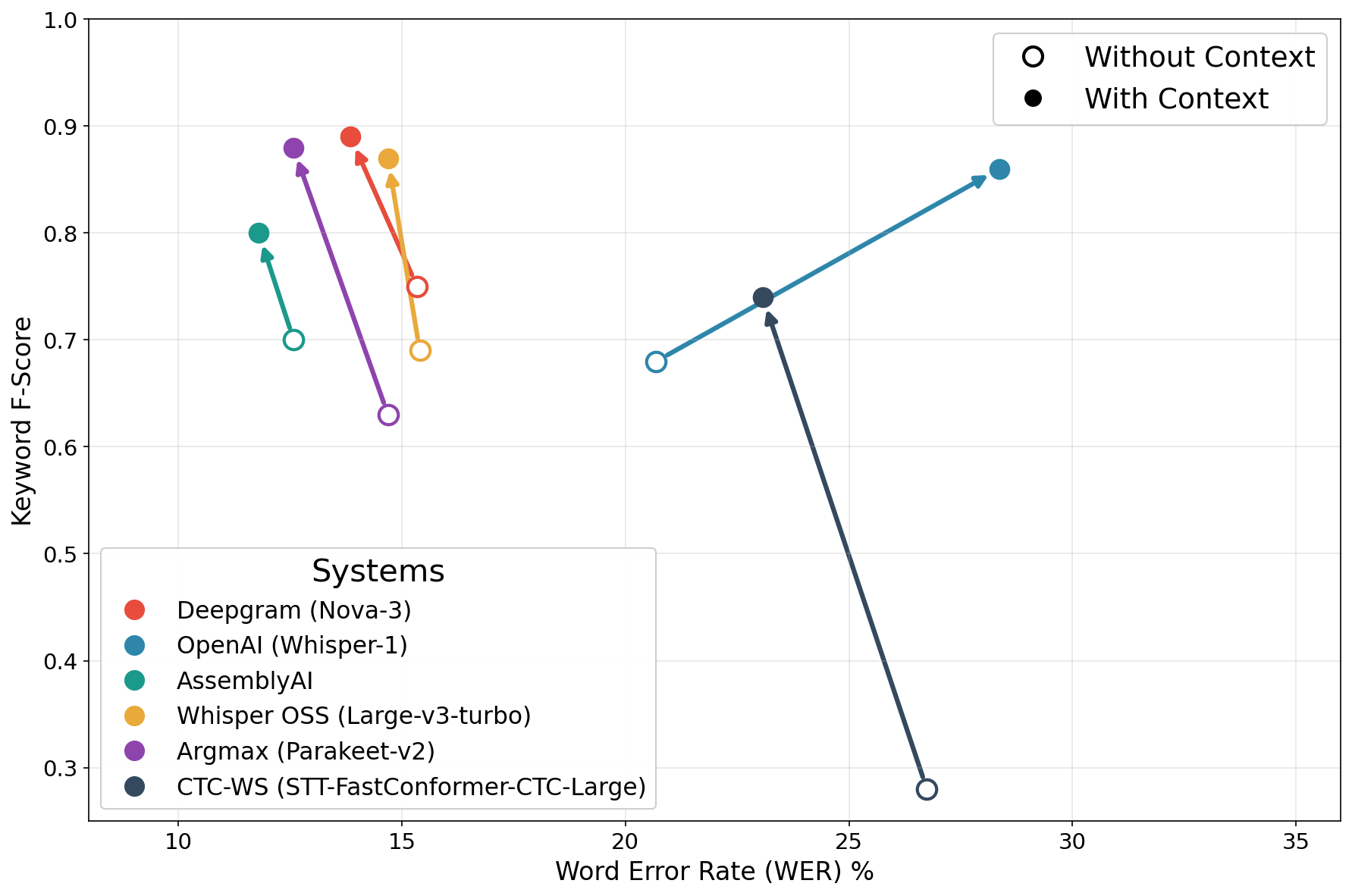}
  \caption{Keyword F-Score vs Word Error Rate comparison across different systems and keyword contexts.}
  \label{fig:fscore_wer}
\end{figure}

\paragraph{Contextual STT methods.}
In practice, two methods dominate deployments and the literature.
\emph{Keyword boosting} integrates a term list into decoding to increase their likelihood when acoustically plausible, spanning shallow-fusion style approaches and GPU-accelerated decoders \cite{ctcws,turbobias,flexctc,xu23d,le21,fox22}.
\emph{Keyword prompting} conditions STT on a keyword list via a textual prompt \cite{whisper,deepgram_prompting,openai_api}.
Recent work also studies \emph{scalability} to long bias lists and mitigations such as ranking/selection of bias terms \cite{hou25}.

\paragraph{Benchmarks and datasets for contextual STT.}
Evaluation for contextual STT remains fragmented.
Many influential contextual-ASR papers rely on \emph{private} or \emph{synthetic} evaluation setups, often constructed by injecting ``rare words'' from a general-domain corpus and adding random distractors.
For example, \cite{le21} evaluates primarily on LibriSpeech with per-utterance synthetic bias lists (rare reference words plus up to thousands of distractors) and additionally reports results on in-house data; the accompanying ``IS21 deep bias'' recipe is publicly released, but the synthetic construction does not reflect naturally occurring, domain-specific entity inventories.
Similarly, streaming and transducer-based contextualization work such as \cite{xu23d} evaluates on LibriSpeech and internal voice-assistant data, limiting cross-paper comparability.
More recent scaling work \cite{hou25} studies large bias lists using LibriSpeech with bias words derived from named entities and the IS21 bias list, again yielding an ad-hoc but controlled setup rather than a domain-realistic benchmark.
Earnings-22 \cite{earnings22} corpora provide a natural public domain where proper nouns are dense and errors on them are high-impact.
ConEC \cite{huang24_conec} augments Earnings-22 with biasing lists derived largely from \emph{external} sources (e.g., slides, earnings releases, and participant metadata), alongside transcript cleanup and sentence-level segmentation, and reports WER-based evaluation on long-form audio.
Separately, Earnings22-Cleaned-AA \cite{earnings22_cleaned_aa,artificialanalysis2026earnings22cleaned} targets reference quality by cleaning transcripts for a very small subset of Earnings-22, but does not introduce contextual vocabularies or contextual evaluation protocols.
Despite this progress, existing public resources still lack a \emph{widely adopted, standardized} benchmark that pairs \emph{context-dense short clips} with \emph{direct} (in-conversation) custom-vocabulary contexts and evaluates both an \emph{idealized} precise-context regime and a \emph{deployment-realistic} distractor regime, enabling apples-to-apples comparison across prompting vs.\ boosting.
\\
\paragraph{Contextual Earnings-22.}\footnotemark[2]
We introduce Contextual Earnings-22, a public dataset built on Earnings-22 \cite{earnings22} that targets the most consequential contextual errors in earnings-call transcription: \textbf{person, company, and product names}.
Each audio segment is paired with realistic custom-vocabulary contexts, and we evaluate two practical scenarios: \textbf{local context} without distractors and \textbf{global context} with realistic distractors.
To support reliable comparison, we manually review and correct the transcripts where needed, addressing earnings-call transcript artifacts at substantially broader coverage than prior cleaned subsets \cite{earnings22_cleaned_aa,artificialanalysis2026earnings22cleaned}.
We build on a previously released open-source evaluation harness \footnotemark[2] by adding keyword boosting evaluation protocols and reproducible baselines.
We set strong baselines spanning the two dominant contextualization families: keyword prompting via widely used STT APIs \cite{whisper,deepgram_prompting,openai_api} and keyword boosting via a scalable CTC-WS pipeline \cite{ctcws}.

\footnotetext[2]{Code and dataset will be released upon acceptance.}
\section{Methodology}
\label{sec:method}
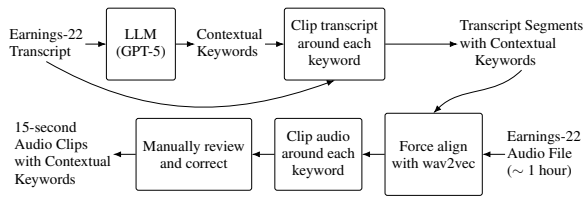
\begin{figure}[!ht]
  \centering
  \resizebox{0.95\columnwidth}{!}{%
  \begin{tikzpicture}[
    font=\Huge,
    >={Latex[scale=1.5]},
    line width=1.2pt,
    node distance=1.2cm,
    every node/.style={align=center},
    box/.style={draw, very thick, rounded corners=4pt, minimum height=42mm, minimum width=40mm, font=\Huge, inner sep=4mm},
    bigbox/.style={draw, very thick, rounded corners=4pt, minimum height=48mm, minimum width=46mm, font=\Huge, inner sep=4mm},
    note/.style={align=left, font=\Huge},
    arrow/.style={->, very thick}
  ]

  \node[note, xshift=2cm] (gt) {Earnings-22\\Transcript};
  \node[box, right=1.2cm of gt] (gpt) {LLM\\(GPT-5)};
  \node[note, right=1.1cm of gpt] (names) {Contextual\\Keywords};
  \node[box, right=1.1cm of names] (cliptr) {Clip transcript\\around each\\keyword};
  \node[note, right=1.1cm of cliptr, xshift=3.1cm] (issnames) {Transcript Segments\\with Contextual\\ \quad Keywords};

  \draw[arrow] (gt) -- (gpt);
  \draw[arrow] (gpt) -- (names);
  \draw[arrow] (names) -- (cliptr);
  \draw[arrow] (cliptr) -- (issnames);
  \draw[arrow] (gt.south) to[out=330, in=200] (cliptr.south);

  \node[bigbox, below=2.5cm of issnames, xshift=-5.1cm] (wav) {Force align\\with wav2vec};
  \draw[arrow] (issnames.south) to[out=225, in=45] (wav.north);

  \node[note, right=1.3cm of wav] (audio) {Earnings-22\\Audio File\\($\sim$ 1 hour)};
  \draw[arrow] (audio) -- (wav);

  \node[box, left=1.3cm of wav] (clip) {Clip audio\\around each\\keyword};
  \draw[arrow] (wav) -- (clip);

  \node[box, left=1.3cm of clip] (mr2) {Manually review\\and correct};
  \draw[arrow] (clip) -- (mr2);

  \node[note, left=1.3cm of mr2] (final) {15-second\\Audio Clips\\with Contextual\\Keywords};
  \draw[arrow] (mr2) -- (final);

  \end{tikzpicture}}
  \caption{Contextual Earnings-22 creation pipeline. Manual review substantially reduced transcript artifacts in the overlapping portion of the dataset: 98.7\% of the samples are free of \emph{inaudible} and \texttt{<unk>} tags, and 29.5\% of clips receive word-level corrections, including spelling fixes as well as word insertions and deletions, affecting 411 words in total.}
  \label{fig:pipeline}
\end{figure}

\subsection{Contextual keyword extraction}
Using Earnings-22 \cite{earnings22} audio--transcript pairs ($\sim$1h per call), we curate samples with contextual keywords focusing on three categories: \textbf{person}, \textbf{company}, and \textbf{product names}.
For each Earnings-22 sample, we run an LLM-based named-entity pass (GPT-5) over the transcript to obtain candidate keywords.
To make the resulting keyword lists stable and evaluation-friendly, we apply deterministic post-processing: (i) de-duplication across surface forms, (ii) punctuation and whitespace normalization, and (iii) filtering of generic strings.
The final \textbf{per-call} keyword inventory defines that call's \emph{global context} list, which naturally varies across calls and includes realistic distractors (Section~\ref{sec:contexts}).

\subsection{Transcript segmentation}
We construct \emph{candidate segments} by locating keyword mentions in the transcript and extracting a local text window (Figure~\ref{fig:pipeline}, top).
Segments are anchored on at least one target keyword but may contain multiple (e.g., a product alongside a company).
For each segment, we record:
(1) segment text,
(2) \emph{local context} (keywords occurring in the segment), and
(3) \emph{global context} (the full call inventory).

\subsection{Forced alignment}
\label{sec:alignment}
To map transcript segments to the long-form audio, we perform forced alignment using a wav2vec-based aligner \cite{wav2vec,wav2vec_og}, obtaining approximate word-level boundaries.
For each keyword mention, we use these boundaries to extract a fixed-length \textbf{15-second} audio window centered on the mention.
We then associate the clip with the corresponding transcript segment and keyword metadata.

\begin{table}[t]
  \captionsetup{justification=raggedright,singlelinecheck=false}
  \caption{Comparison of transcripts before (left column) \\
  and after (right column) manual correction.}
  \label{tab:corrections}
  \scriptsize
  \begin{tabular}{p{3.5cm}p{3.5cm}}
  \toprule
  \textbf{Original} & \textbf{Corrected}  \\
  \midrule
  obviously also the restrictions that are still in place and that also affect our financial services activity. Okay. Very clear, thank you everyone. Okay. Uh, \textcolor{red}{inaudible}, Paolo, can you, uh, shed some light, uh, on the question that \textcolor{red}{Artur}
  & obviously also the restrictions that are still in place and that also affect our financial services activity. Okay. Very clear, thank you everyone. Okay. Uh, \textcolor{blue}{Rui} Paulo, can you, uh, shed some light, uh, on the question that \textcolor{blue}{Arthur} \\
  \hline
  \\
  Okay. Thanks, \textcolor{red}{Panto}. Appreciate that. Your next question comes from the line of Nico \textcolor{red}{Margaroni's} of BRI \textcolor{red}{Benelexa}. Please ask your question
  & Okay. Thanks \textcolor{blue}{Mark}. Appreciate that. Your next question comes from the line of Nico \textcolor{blue}{Margaronis} of BRI \textcolor{blue}{Danareksa}. Please ask your question \\
  \hline
  \\
  call is our CEO Damian Scokin, who \textcolor{red}{is inaudible} on our specific priorities. \textcolor{red}{Lastly}, our \textcolor{red}{CSO} will then discuss the third quarter's financial return. After that we'll all- we'll open
  & call is our CEO Damian Scokin, who \textcolor{blue}{will give you an overview of the third quarter and} on our specific priorities. \textcolor{blue}{Alberto López-Gaffney}, our \textcolor{blue}{CFO} will then discuss the third quarter's financial return. After that we'll all- we'll open \\
  \hline
  \\
  the first one is, um, Jonathan \textcolor{red}{unk} at, uh, \textcolor{red}{inaudible} Investments, um, probably for Gus. Does the HBC other net income come back in future years, or is it permanently gone? Thanks, Jonathan. Um
  & the first one is, um, Jonathan \textcolor{blue}{du Toit} at, uh, \textcolor{blue}{Oyster Catcher} Investments, um, probably for Gus. Does the HBC other net income come back in future years, or is it permanently gone? Thanks, Jonathan. Um \\
  \bottomrule
  \end{tabular}
\end{table}

\subsection{Manual review and correction}
To prevent errors from confounding evaluation, we manually review each candidate.
Review focuses on:
(i) \textbf{transcript fidelity} (text matches audio),
(ii) \textbf{keyword validity} (targets are actually spoken and correctly typed), and
(iii) \textbf{alignment sanity} (audio corresponds to the intended text).
We correct artifacts including mis-heard names, casing/punctuation inconsistencies, acronym formatting, and boundary errors for multi-word entities. Table~\ref{tab:corrections} shows representative corrections.

\subsection{Context construction and release}
\label{sec:contexts}
We support two context regimes mirroring product settings:\\
\textbf{Local context} contains only keywords that appear in the target clip/segment, isolating a system's ability to leverage relevant context when it is precise.\\
\textbf{Global context} is derived from the broader call-level inventory(i.e., keywords extracted from the full one-hour source audio from which the clip is segmented), which naturally includes keywords not spoken in the clip (distractors), capturing the precision--recall trade-offs that arise when users provide long custom vocabularies in real deployments.\\
The released benchmark includes these context lists, the 15-second audio clips, reviewed transcripts, and an open-source evaluation harness to ensure reproducibility.
Table~\ref{tab:dataset_stats} summarizes the dataset statistics.

\begin{table}[ht!]
  \captionsetup{justification=raggedright,singlelinecheck=false}
  \caption{Statistics of the Contextual Earnings-22 dataset. A 55-file subset yields 760 context-dense 15-second samples, split into validation (for hyperparameter tuning) and test (for benchmarking) sets at the source-audio level to prevent leakage.}
  \label{tab:dataset_stats}
  \centering
  \footnotesize
  \begin{tabular}{lcc}
  \toprule
  & \textbf{Validation} & \textbf{Test} \\
  \midrule
  Samples              & 130  & 630   \\
  Total keyword instances & 248  & 1{,}259 \\
  Unique keywords      & 134  & 738   \\
  Source files          & 9    & 46    \\
  Source audio duration (h) & 9.14 & 58.47 \\
  \bottomrule
  \end{tabular}
\end{table}

\begin{table*}[ht]
  \caption{Illustrative failure examples comparing STT outputs with and without context. In the predictions, \textcolor{blue}{blue} denotes correctly predicted keywords (TP), and \textcolor{red}{red} denotes incorrectly predicted keywords (FP).}
  \label{tab:qualitative_examples}
  \centering
  \fontsize{6pt}{7.2pt}\selectfont
  \begin{tabular}{p{1cm}p{2.5cm}p{3.5cm}p{3.5cm}p{3.5cm}}
  \toprule
  \textbf{System} &\textbf{Contextual Keywords} & \textbf{Ground Truth} & \textbf{Prediction without Context} & \textbf{Prediction with Context} \\
  \midrule
  Argmax
  & Dan, Johan, Julien, Exane, Quenouille
  & So, um, next one online should be Julien Quenouille from Exane. Uh, Julien you should be live. Hello. Good morning Dan. Good morning Johan thanks, thanks for taking my questions. I have two. Uh, the first one a- and sorry if you
  & So um next one online should be \textcolor{blue}{Julien} Dunnoir, Ray Sain. Um Julian uh should be live Hello good morning Don good morning Yoran thanks thanks for taking my questions. I have two. Uh the first one and sorry if you 're
  & So um next one online should be \textcolor{blue}{Julien} \textcolor{red}{Dan} Ray \textcolor{blue}{Exane} Um \textcolor{blue}{Julien} uh should be live Hello good morning \textcolor{blue}{Dan} good morning \textcolor{red}{Julien} thanks thanks for taking my questions. I have two. Uh the first one and sorry if you 're\\
  \hline
  \\
  Deepgram
  &  Dan, Johan, Julien, Exane, Quenouille
  &  So, um, next one online should be Julien Quenouille from Exane. Uh, Julien you should be live. Hello. Good morning Dan. Good morning Johan thanks, thanks for taking my questions. I have two. Uh, the first one a- and sorry if you
  &  So next one on line should be \textcolor{blue}{Julien} Dunois, DUMOULIN Rex Hain. \textcolor{red}{Julien}, should be live. SMITH:\rbrack \quad Hello, good morning, \textcolor{blue}{Dan}. Good morning, Jorgen. Thanks for taking my questions. I have two. The first one, and sorry if you
  &  So next one on line should be \textcolor{blue}{Julien} Dumoulin, DUMOULIN Rex Hain. \textcolor{red}{Julien}, should be live. SMITH:\rbrack \quad Hello, good morning \textcolor{blue}{Dan}, good morning Jorgen. Thanks for taking my questions. I have two. The first one, and sorry if you \\
  \hline
  \\
  OpenAI
  & Nico, Margaronis, BRI Danareksa
  & Thanks Mark. Appreciate that. Your next question comes from the line of Nico Margaronis of BRI Danareksa. Please ask your question
  & Thanks a lot appreciate that Your next question comes from the line of Meiko \textcolor{blue}{Margaronis} of \textcolor{red}{BRI Danarexa} please ask your question
  & Thanks a lot appreciate that \\
  \hline
  \\
  Whisper OSS
  & Sify
  & across the country. A 6\% and 11\% increase respectively over the same quarter last year. As part of this digital experience project, Sify completed full automation of service assurance
  & across the country. A 6 \% and 11 \% increased respectively over the same quarter last year. As part of his digital experience project, SIFI completed full automation of service assurance.
  & As part of his digital experience project, \textcolor{blue}{Sify}, \textcolor{red}{Sify}, \textcolor{red}{Sify}, \textcolor{red}{Sify}, \textcolor{red}{Sify}, \textcolor{red}{Sify}, \textcolor{red}{Sify}, \textcolor{red}{Sify}, \textcolor{red}{Sify}, \textcolor{red}{Sify}, \textcolor{red}{Sify}, \textcolor{red}{Sify}, \textcolor{red}{Sify}, \textcolor{red}{Sify}, \textcolor{red}{Sify} and \textcolor{red}{Sify}. \\
  \hline
  \\
  OpenAI
  & Citi, Samarth, Agarwal
  & and you read them out, uh… Thank you. Sorry. And the first one is from Samarth Agarwal of Citi, thanks for taking my questions too, for me, how much flexibility do you have to pass on input costs to customers if
  & and you read them out Thank you Sorry And the first one is from \textcolor{blue}{Samarth Agarwal} of \textcolor{blue}{Citi} Thanks for taking my questions Two for me how much flexibility do you have to pass on input costs to customers
  & Låt oss läsa dem ut Tack Den första är från \textcolor{blue}{Samarth Agarwal} från \textcolor{blue}{Citi} Tack för att du tog mina frågor Två från mig Hur mycket flexibilitet måste man lägga på kostnaderna till kunderna \\
  \bottomrule
  \end{tabular}
  \end{table*}

  \section{Metrics}
  \label{sec:metrics}

  We report two complementary metrics: Word Error Rate (WER) and \textbf{keyword-centric} metrics measuring contextual word recognition quality.\\
  \textbf{WER.} We report standard WER between the STT hypothesis and the reference transcript for each clip.\\
  \textbf{Keyword Precision/Recall/F-score.} Recent STT systems can have very similar aggregate WER on common benchmarks, while still differing substantially in whether they correctly recognize  \emph{context-defined} words that determine transcript usefulness in practice. For measuring this, we report keyword-centric metrics that \emph{isolate} performance on the provided contextual keyword list.
  A keyword is a \textbf{True Positive (TP)} if and only if it matches the reference text and aligned location exactly, where alignment is computed via minimum edit distance \cite{levenshtein,texterrors}. Otherwise, it is a \textbf{False Negative (FN)} (if in reference but not hypothesis) or a \textbf{False Positive (FP)} (if in hypothesis but not reference, including misaligned correct text). Precision, recall, and F-score are then computed \emph{per sample} from these TP/FP/FN counts. \\
\section{Results}
\label{sec:results}

We evaluate six STT systems under \textbf{no}, \textbf{local}, and \textbf{global} context, reporting WER and keyword F-score (precision/recall). 
\subsection{Benchmarked systems}
All systems are benchmarked reproducibly using the same open-source evaluation harness \footnotemark[2]
\begin{itemize}
  \item \textbf{Deepgram (Nova-3)} \cite{deepgram_prompting}: a commercial STT API with keyword prompting support, representing a commercial-scale keyword prompting baseline.
  \item \textbf{OpenAI (Whisper-1)} \cite{openai_api}: OpenAI's official STT API, based on the Whisper Large v2 architecture. Contextual conditioning is applied via the optional prompt parameter, representing a commercial-scale keyword prompting baseline.
  \item \textbf{AssemblyAI} \cite{assemblyai}: a commercial STT API with keyword prompting support, representing another commercial-scale keyword prompting baseline.
  \item \textbf{Whisper OSS (Large-v3-turbo)} \cite{whisper,whisper_oss}: OpenAI's official open-source Whisper implementation. Contextual conditioning is applied via the prompt parameter exposed in the official repository.
  \item \textbf{CTC-WS (STT-FastConformer-CTC-Large)}
  \cite{ctcws}:
  the default CTC-WS configuration using STT-FastConformer-CTC-Large which is the default model in \cite{ctcws} and its open-source implementation. Hyperparameters are calibrated on the contextual Earnings-22 validation split to achieve the best results.
  \item \textbf{Argmax (Parakeet-v2 + CTC-WS)}
  \cite{ctcws,parakeetv2-hf,parakeet-v3,whisperkit25}:
  a state-of-the-art open STT model paired with CTC-WS, a CTC-based keyword boosting method \cite{ctcws}.
  The official CTC-WS implementation is used with two optional CTC backbones for English and multilingual settings \cite{canary-ctc,parakeet-v3,parakeet-ctc}.
  Unlike the original setup \cite{ctcws}, inference follows two separate paths: STT is performed via Parakeet-TDT-0.6B-v2 \cite{parakeetv2-hf}, and CTC keyword scoring is computed using the corresponding CTC backbone \cite{parakeet-ctc}.
  The two inference paths are combined using a slightly modified integration strategy.
  Hyperparameters are calibrated on the contextual Earnings-22 validation split to achieve the best results.
  \end{itemize}
\begin{figure}[!ht]
  \centering
  \includegraphics[width=\linewidth]{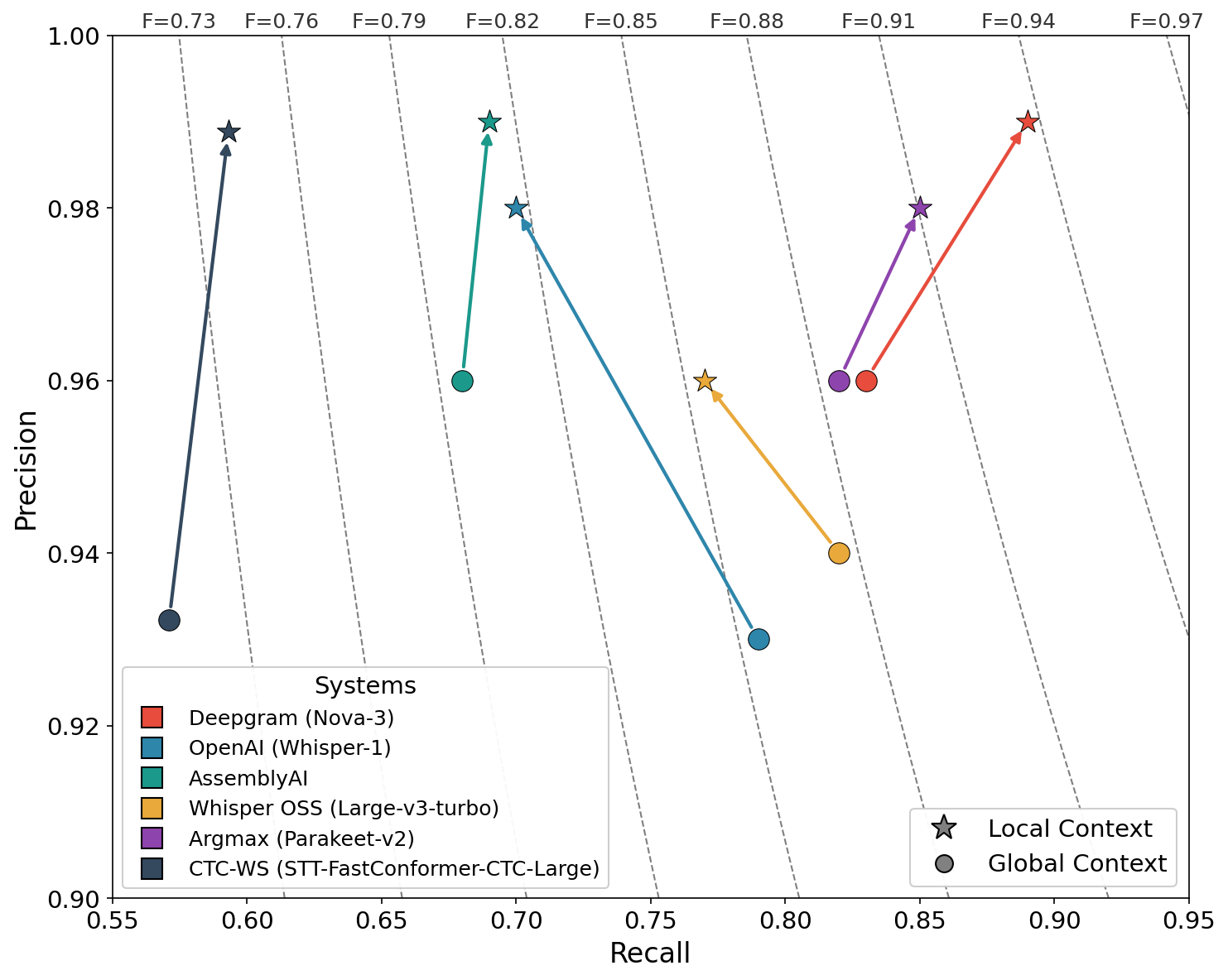}
  \caption{Precision and Recall comparison across different systems and keyword contexts. Dashed lines represent F-score iso-curves. See supplementary Tables~S1 and~S2 for exact numbers.}
  \label{fig:keyword_comparison}
\end{figure}
\subsection{Context conditioning improves transcript quality}
STT systems with effective contextual conditioning  capability should achieve higher keyword F-score than in the no-context setting while maintaining comparable WER. When conditioning does not introduce artifacts (e.g., hallucinations, high false positive rate), these keyword corrections should also translate into lower WER. Figure~\ref{fig:fscore_wer} shows keyword F-score versus WER for the benchmarked systems with and without context. Across all systems, providing context yields a increase in \textbf{keyword F-score}, indicating that contextual conditioning substantially improves recognition of contextual terms. In contrast, changes in \textbf{WER} are smaller and less consistent: some systems improve WER modestly, while others show little change or even worse WER despite markedly better keyword F-score.\\
The consistent pattern in Figure~\ref{fig:fscore_wer} is that context shifts operating points upward in keyword F-score for every system, but WER changes vary by system; for OpenAI, WER increases under context in our setup. This suggests that contextual capability of different systems have different rate of artifacts, motivating keyword-centric evaluation alongside WER. Some illustrative examples of artifacts are shown in Table~\ref{tab:qualitative_examples}.
\subsection{Local vs global context trade-offs}
Figure~\ref{fig:keyword_comparison} compares keyword precision and recall under global context and local context. Two consistent observations emerge.\\
\textbf{Local context is systematically easier.} Most systems move toward higher F-score iso-curves under local context which can be seen in Figure~\ref{fig:keyword_comparison}, reflecting higher precision and/or recall when the provided context is concise and relevant. This is expected: without distractors the risk of inserting non-spoken contextual words is reduced.
\\
\textbf{Global context primarily stresses precision.} With global context, systems face a realistic deployment regime where the context list contains many plausible-but-absent terms. Even when recall remains strong, precision can drop due to \emph{distractor-induced false positives}. This is visible in Figure~\ref{fig:keyword_comparison} as global-context points tending to lie on lower iso-F curves than local-context points for the same system.
\\
Interestingly, the magnitude and direction of the local--global shift differs by system family. Several systems exhibit large precision gains under local context with smaller recall changes, consistent with global-context distractors being the dominant source of error. Other systems exhibit a more pronounced recall shift between regimes, suggesting sensitivity to context formatting or the mechanism used to incorporate the keyword list. These differences motivate reporting both regimes: \textbf{local context} measures the ability to exploit relevant context, while \textbf{global context} measures robustness to realistic, noisy context lists.
\section{Discussion \& Conclusion}

\paragraph{Qualitative error modes.}
Table~\ref{tab:qualitative_examples} highlights representative behaviors that help interpret the precision--recall trade-offs observed under local and global context.
First, \textbf{context resolves near-miss confusions for rare names}: without context, proper nouns are often substituted with phonetically similar strings or fragmented into partial matches; providing the correct vocabulary can convert these near-misses into correct keyword hits, improving recall.
Second, \textbf{global-context distractors can induce false insertions}: when the context list contains many plausible-but-absent names, some systems insert context terms that are not supported by the audio, reducing precision.
Third, \textbf{prompting can change STT behavior beyond keyword insertion}. In addition to keyword-level effects, we observe prompt-induced artifacts that are visible in Table~\ref{tab:qualitative_examples}, including (i) \emph{hallucinations} where provided context words are inserted despite not being spoken, (ii) \emph{partial or empty outputs} where the model deviates from its no-context STT behavior, and (iii) occasional \emph{language switching} when the keyword list perturbs the decoding trajectory. Taken together, these behaviors motivate reporting \emph{both} keyword-centric metrics and WER as complementary signals: keyword metrics capture recognition of the provided custom vocabulary, while WER captures broader side effects that may degrade overall transcript quality even when keyword recognition improves (and vice versa).

\paragraph{Conclusion.}
Overall, Contextual Earnings-22 provides a standardized, public benchmark for contextual STT that pairs short earnings-call clips with realistic custom-vocabulary contexts and evaluates both an idealized regime (local context) and a deployment-realistic regime (global context with distractors). Our baseline results show substantial improvements in contextual term recognition across both keyword prompting and keyword boosting approaches, while also revealing that \textbf{robustness to distractors} remains a key differentiator between systems. We release the audio clips, reviewed transcripts, context lists, and an open-source evaluation harness to enable reproducible comparison and accelerate progress on contextual speech recognition.

\bibliographystyle{IEEEtran}
\bibliography{main}

@misc{earnings22,
  title        = {Earnings-22: A Practical Benchmark for Accents in the Wild},
  author       = {Del Rio, Miguel and Ha, Peter and McNamara, Quinten and Miller, Corey and Chandra, Shipra},
  year         = {2022},
  howpublished = {arXiv preprint arXiv:2203.15591},
  url          = {https://arxiv.org/abs/2203.15591}
}

@inproceedings{ctcws,
  title={Fast Context-Biasing for CTC and Transducer ASR Models with CTC-based Word Spotter},
  author={Andrusenko, Andrei and Laptev, Aleksandr and Bataev, Vladimir and Lavrukhin, Vitaly and Ginsburg, Boris},
  booktitle={Proc. Interspeech},
  year={2024},
  doi={10.21437/Interspeech.2024-1002}
}

@article{turbobias,
  title={TurboBias: Universal ASR Context-Biasing Powered by GPU-Accelerated Phrase-Boosting Tree},
  author={Andrusenko, Andrei and Bataev, Vladimir and Grigoryan, Lilit and Lavrukhin, Vitaly and Ginsburg, Boris},
  journal={arXiv preprint},
  volume={abs/2508.07014},
  year={2025},
  url={https://arxiv.org/abs/2508.07014}
}

@article{flexctc,
  title={FlexCTC: GPU-Powered CTC Beam Decoding with Advanced Contextual Abilities},
  author={Grigoryan, Lilit and Bataev, Vladimir and Karpov, Nikolay and Andrusenko, Andrei and Lavrukhin, Vitaly and Ginsburg, Boris},
  journal={arXiv preprint},
  volume={abs/2508.07315},
  year={2025},
  url={https://arxiv.org/abs/2508.07315}
}

@inproceedings{xu23d,
  title={Adaptive Contextual Biasing for Transducer-Based Streaming Speech Recognition},
  author={Xu, Tianyi and Yang, Zhanheng and Huang, Kaixun and Guo, Pengcheng and Zhang, Ao and Li, Biao and Chen, Changru and Li, Chao and Xie, Lei},
  booktitle={Proc. Interspeech},
  year={2023},
  pages={1668--1672},
  doi={10.21437/Interspeech.2023-884}
}

@inproceedings{le21,
  title={Contextualized Streaming End-to-End Speech Recognition with Trie-Based Deep Biasing and Shallow Fusion},
  author={Le, Duc and Jain, Mahaveer and Keren, Gil and Kim, Suyoun and Shi, Yangyang and Mahadeokar, Jay and Chan, Julian and Shangguan, Yuan and Fuegen, Christian and Kalinli, Ozlem and Saraf, Yatharth and Seltzer, Michael L.},
  booktitle={Proc. Interspeech},
  year={2021},
  pages={1772--1776},
  doi={10.21437/Interspeech.2021-1566}
}

@inproceedings{fox22,
  title={Improving Contextual Recognition of Rare Words with an Alternate Spelling Prediction Model},
  author={Fox, Jennifer Drexler and Delworth, Natalie},
  booktitle={Proc. Interspeech},
  year={2022},
  pages={3914--3918},
  doi={10.21437/Interspeech.2022-10991}
}

@inproceedings{hou25,
  title={Ranking and Selection of Bias Words for Contextual Bias Speech Recognition},
  author={Hou, Haoxiang and Gong, Xun and Zhang, Wangyou and Wang, Wei and Qian, Yanmin},
  booktitle={Proc. Interspeech},
  year={2025},
  pages={5183--5187},
  doi={10.21437/Interspeech.2025-646}
}

@inproceedings{whisper,
  title={Robust Speech Recognition via Large-Scale Weak Supervision},
  author={Radford, Alec and Kim, Jong Wook and Xu, Tao and Brockman, Greg and McLeavey, Christine and Sutskever, Ilya},
  booktitle={Proceedings of the 40th International Conference on Machine Learning},
  series={Proceedings of Machine Learning Research},
  volume={202},
  pages={28492--28518},
  year={2023},
  publisher={PMLR},
}

@misc{whisper_oss,
  title={Whisper: Official OpenAI Open-Source Repository},
  author={{OpenAI}},
  howpublished={\url{https://github.com/openai/whisper}},
  year={2023}
}

@misc{openai_api,
  title        = {Speech to Text Guide},
  author       = {{OpenAI}},
  howpublished = {OpenAI API Documentation},
  year         = {2025},
  url          = {https://developers.openai.com/api/docs/guides/speech-to-text/}
}

@misc{deepgram_prompting,
  title={Deepgram Keyterm Prompting},
  author={{Deepgram}},
  howpublished={\url{https://developers.deepgram.com/docs/keyterm}},
  year={2024},
  note={API feature documentation (no formal published paper)}
}

@article{parakeet-v3,
  title={Canary-1B-v2 \& Parakeet-TDT-0.6B-v3: Efficient and High-Performance Models for Multilingual ASR and AST},
  author={Sekoyan, Monica and Koluguri, Nithin Rao and Tadevosyan, Nune and Zelasko, Piotr and Bartley, Travis and Karpov, Nick and Balam, Jagadeesh and Ginsburg, Boris},
  journal={arXiv preprint arXiv:2509.14128},
  year={2025},
  url={https://arxiv.org/abs/2509.14128}
}

@misc{parakeetv2-hf,
  title        = {{Parakeet-TDT-0.6B V2}: Automatic Speech Recognition Model},
  author       = {{NVIDIA}},
  howpublished = {Hugging Face model card},
  year         = {2025},
  url          = {https://huggingface.co/nvidia/parakeet-tdt-0.6b-v2}
}

@misc{parakeet-ctc,
  title        = {{Parakeet-TDT\_CTC-110M}: English Automatic Speech Recognition Model},
  author       = {{NVIDIA \& Suno.ai}},
  howpublished = {Hugging Face Model Card},
  year         = {2025},
  url          = {https://huggingface.co/nvidia/parakeet-tdt_ctc-110m}
}

@misc{canary-ctc,
  title        = {{Canary-1B-v2}: Multilingual ASR and AST Model},
  author       = {{NVIDIA}},
  howpublished = {Hugging Face Model Card},
  year         = {2025},
  url          = {https://huggingface.co/nvidia/canary-1b-v2}
}

@misc{assemblyai,
  title        = {Keyterms Prompting Documentation},
  author       = {{AssemblyAI}},
  howpublished = {AssemblyAI Documentation},
  year         = {2024},
  url          = {https://www.assemblyai.com/docs/pre-recorded-audio/keyterms-prompting}
}

@inproceedings{whisperkit25,
  title={WhisperKit: On-device Real-time ASR with Billion-Scale Transformers},
  author={Durmus, Berkin and Okan, Arda and Pacheco, Eduardo and Nagengast, Zach and Orhon, Atila},
  booktitle={Proceedings of the Tiny Titans: The Next Wave of On-Device Learning for Foundation Models (TTODLer-FM) Workshop, ICML 2025},
  year={2025},
  month={July},
  address={Vancouver, Canada},
  note={Presented at TTODLer-FM @ ICML 2025},
  url={https://openreview.net/forum?id=6lC3MPFbVg}
}

@misc{artificialanalysis2026earnings22cleaned,
  title  = {Earnings22-Cleaned-AA: Cleaned Ground Truth Transcripts for Earnings22 English Test Set},
  author = {Artificial Analysis},
  year   = {2026},
  url    = {https://artificialanalysis.ai/articles/aa-wer-v2}
}

@misc{earnings22_cleaned_aa,
  title        = {Earnings22-Cleaned-AA},
  author       = {{Artificial Analysis}},
  year         = {2026},
  howpublished = {Hugging Face Datasets},
  url          = {https://huggingface.co/datasets/ArtificialAnalysis/Earnings22-Cleaned-AA},
}

@inproceedings{huang24_conec,
  title     = "{C}on{EC}: Earnings Call Dataset with Real-world Contexts for Benchmarking Contextual Speech Recognition",
  author    = "Huang, Ruizhe  and
               Yarmohammadi, Mahsa  and
               Trmal, Jan  and
               Liu, Jing  and
               Raj, Desh  and
               Garcia, Leibny Paola  and
               Ivanov, Alexei  and
               Ehlen, Patrick  and
               Yu, Mingzhi  and
               Rastrow, Ariya  and
               Povey, Dan  and
               Khudanpur, Sanjeev",
  booktitle = "Proceedings of the 2024 Joint International Conference on Computational Linguistics, Language Resources and Evaluation (LREC-COLING 2024)",
  year      = "2024",
  address   = "Torino, Italia",
  publisher = "ELRA and ICCL",
  url       = "https://aclanthology.org/2024.lrec-main.328/",
  pages     = "3700--3706"
}

@misc{wav2vec,
  author       = {Moto Hira},
  title        = {Forced Alignment with Wav2Vec2},
  howpublished = {\url{https://docs.pytorch.org/audio/stable/tutorials/forced_alignment_tutorial.html}},
  year         = {2025},
}

@inproceedings{wav2vec_og,
  title     = {{wav2vec: Unsupervised Pre-Training for Speech Recognition}},
  author    = {Steffen Schneider and Alexei Baevski and Ronan Collobert and Michael Auli},
  year      = {2019},
  booktitle = {{Interspeech 2019}},
  pages     = {3465--3469},
  doi       = {10.21437/Interspeech.2019-1873},
  issn      = {2958-1796},
}

@article{levenshtein,
  title={Binary codes capable of correcting deletions, insertions, and reversals},
  author={Levenshtein, Vladimir I},
  journal={Soviet Physics Doklady},
  volume={10},
  pages={707--710},
  year={1966}
}

@software{texterrors,
  author = {Braun, Ruben A.},
  title = {texterrors: Text alignment and error analysis in Python},
  year = {2023},
  url = {https://github.com/RuABraun/texterrors},
}

@misc{hf_open_asr_leaderboard,
  title  = {Open ASR Leaderboard},
  author = {Srivastav, Vaibhav and Majumdar, Somshubra and Koluguri, Nithin and Moumen, Adel and Gandhi, Sanchit and Hugging Face Audio Team},
  year   = {2023},
  url    = {https://huggingface.co/spaces/hf-audio/open_asr_leaderboard},
}

\end{document}